\pdfoutput=1

\documentclass[11pt]{article}

\usepackage[final]{acl}

\usepackage{times}
\usepackage{latexsym}
\usepackage{arydshln} 

\usepackage[T1]{fontenc}

\usepackage[utf8]{inputenc}

\usepackage{microtype}

\usepackage{inconsolata}

\usepackage{graphicx}

\usepackage{arabtex}
\usepackage{utf8}
\usepackage[T1]{fontenc}

\newcommand{\hide}[1]{}

\newcommand{\AYN}{{E}}

\newcommand{\SHADDA}{{$\sim$}}

\usepackage{booktabs} 
 
\usepackage{makecell}

\title{Lemmatization as a Classification Task:\\ Results from Arabic across Multiple Genres
}

\newcommand*{\authormark}[1][*]{\textsuperscript{#1}}

\author{
  Mostafa Saeed\authormark[1],
  Nizar Habash\authormark[1] \\
  {\normalfont Computational Approaches to Modeling Language (CAMeL) Lab} \\
  {\normalfont\authormark[1]New York University Abu Dhabi} \\
  {\normalfont\texttt{\{mostafa.saeed,nizar.habash\}@nyu.edu}}
}

\begin{document}
\setcode{utf8}
\maketitle
\begin{abstract}
Lemmatization is crucial for NLP tasks in morphologically rich languages with ambiguous orthography like Arabic, but existing tools face challenges due to inconsistent standards and limited genre coverage. This paper introduces two novel approaches that frame lemmatization as classification into a Lemma-POS-Gloss (LPG) tagset, leveraging machine translation and semantic clustering. We also present a new Arabic lemmatization test set covering diverse genres, standardized alongside existing datasets. We evaluate character-level sequence-to-sequence models, which perform competitively and offer complementary value, but are limited to lemma prediction (not LPG) and prone to hallucinating implausible forms. Our results show that classification and clustering yield more robust, interpretable outputs, setting new benchmarks for Arabic lemmatization.

\end{list} 
\end{abstract}


\section{Introduction}
Lemmatization is the process of mapping a word to a base form that abstracts away from its inflectional variants.  Lemmatization has played an important enabling technology role in many NLP applications, including machine translation 
\cite{conforti2018neural}, information retrieval \cite{semmar2006deep}, parsing \cite{seddah-etal-2010-lemmatization}, text classification \cite{abdelrahman2021otrouha} and summarization \cite{el2014lemma}. 
Despite the shift toward large language models, lemmatization remains essential for tasks involving morphologically rich languages and  requiring interpretability, such as readability assessment \cite{AlKhalil:2018:leveled, liberato-etal-2024-strategies} or automated error detection \cite{belkebir-habash-2021-automatic}.

 \begin{table}[t]
     \centering
     \includegraphics[width=1\linewidth]{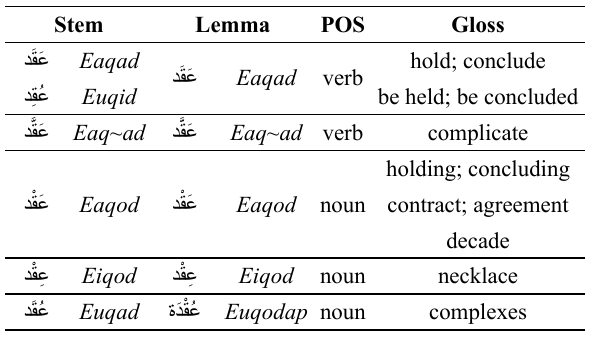}
     \caption{Eight possible Lemma-POS-Gloss analyses for the Arabic word \<عقد> \textit{{\AYN}qd}. Transliteration is in \newcite{Buckwalter:2002:buckwalter}'s scheme.}
     \label{fig:example}\vspace{-10pt}
 \end{table}
 
Lemmatization is especially challenging in morphologically rich languages like Arabic due to complex morphology and optional diacritics. Table~\ref{fig:example} presents multiple out-of-context analyses of a single word, varying in diacritization, lemma, POS, and English gloss (as a proxy for sense).

Previous lemmatization approaches rely on morphological analyzers and ranking models \cite{Roth:2008:arabic}, sequence-to-sequence (seq2seq) generation \cite{Bergmanis:2018:context, zalmout-habash-2020-joint}, or edit-based tagging \cite{gesmundo-samardzic-2012-lemmatisation, Kondratyuk:2018:lemmatag}. However, these methods often focus only on the lemma form, lack generalization across domains, and rely on narrow lexical resources or genre-specific training data.
In this work, we propose a broader and more interpretable framing of lemmatization as classification into a rich Lemma-POS-Gloss (LPG) tagset.
 Our contributions are:

\noindent\textbf{First}, we introduce two novel approaches that classify into LPG labels, one leveraging machine translation of source sentences and dictionary glosses, and the other using LPG semantic clustering.

\noindent\textbf{Second}, we present a new multi-genre Arabic lemmatization test set, covering underexplored domains such as novels and children's stories.


Our experiments demonstrate that LPG-based classification and clustering approaches outperform prior systems that resolve most morphosyntactic ambiguity \cite{inoue2021morphosyntactic}, offering superior accuracy and robustness. We also evaluate character-level seq2seq models, which perform competitively and provide complementary benefits, but are limited to lemma-only (not LPG) prediction and often hallucinate implausible forms. Hybrid models that combine seq2seq and classification techniques further boost performance.
All code, models, and annotations will be released to support continued research in Arabic lemmatization.

\hide{ \begin{itemize}
     \item \textbf{Enhanced lemmatization process:} This study enhances lemmatization by treating it as a classification and grouping problem and incorporating English senses as auxiliary features for improved semantic context.
     \item \textbf{Redefining lemmatization evaluation metrics:} A multi-dimensional evaluation framework is proposed, covering lemma-only, lemma + POS, and lemma + POS + stemgloss for a comprehensive lemmatization assessment.
     \item \textbf{Introducing a new lematization benchmark dataset}
 \end{itemize}


 
}

The paper is organized as follows: \S\ref{ling-back} covers linguistic background, \S\ref{rel-work} reviews related work, \S\ref{data} describes the dataset, \S\ref{approach} outlines our methods, and \S\ref{eval} presents the evaluation results.

\begin{table}[t]
\centering
\tabcolsep3pt
\begin{tabular}{lcccc}
\toprule
\textbf{Unique Avg} & \textbf{All} & \textbf{Top} &  \textbf{Ambig$\downarrow$} & \textbf{Recall} \\
\midrule
\textbf{Analyses} & 15.5 & 1.3 &  91.3\%  & \\
\textbf{LPG}      & 2.7  & 1.3 &  52.8\% & 96.2\% \\
\textbf{LP}       & 2.5  & 1.2 &  52.6\% & 98.8\% \\
\textbf{L}        & 2.0  & 1.2 &  42.4\% & 99.6\% \\
\bottomrule
\end{tabular}
\caption{Avg \# of unique entries of CAMeL Tools analyzer and disambiguator on the ATB Dev set in terms of full morphological analyses, Lemmas (\textbf{L}), Part-of-Speech (\textbf{P}) and Gloss (\textbf{G}) combinations. \textbf{All} refers to all returned unique values per word; and \textbf{Top} refers to all remaining values after filtering with the POS Tagger. \textbf{Ambig$\downarrow$} shows the effect of the POS Tagger. \textbf{Recall} shows the maximal potential accuracy for each representation combination. }
    \label{fig:ambig_analysis}
\end{table}


\section{Linguistic Background}
\label{ling-back}
%

Arabic is morphologically and orthographically rich, with optional diacritics and multiple word forms contributing to ambiguity in both meaning and structure. Previous research has focused on lemma alone (L) or lemma with POS (LP), but none have examined the more complex Lemma, POS, and Gloss (LPG). This study aims to fill this gap while evaluating simpler variations for completeness.

We use the CAMeL Tools analyzer-and-disambiguator system as our baseline \cite{inoue2021morphosyntactic, obeid-etal-2020-camel}, which returns a set of ranked   morphological analyses per word, including gender, number, clitics, POS, and 37 other features. While this helps resolve many morphosyntactic ambiguities, it does not fully disambiguate lemma or sense, which are often given the same rank. For instance, for
\<بعقدها> \textit{bEqdhA}, the model correctly rules out a verbal interpretation, but ambiguity remains among nominal readings such as ‘in her contract,’ ‘necklace,’ or ‘complexes’ (see Table~\ref{fig:example}).


As shown in Table~\ref{fig:ambig_analysis}, the analyzer initially produced an average of 15 analyses per word on the dev set. Restricting to the top-ranked disambiguator output reduced this to 1.3 analyses per word on average, a 91.3\% reduction via morphosyntactic feature tagging. For LPG selection, ambiguity was reduced by 52.8\% but capped at 96.2\% recall. The LPG space remains significantly more ambiguous than LP or L alone, making full disambiguation substantially more challenging.

\begin{table}[t]
    \centering
    \includegraphics[width=\linewidth]{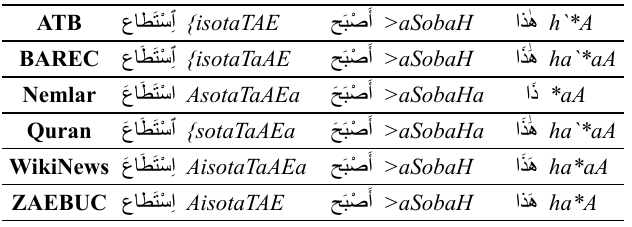}
    \caption{Examples highlighting differences in lemma representations across the data sets we synchronize.}
    \label{fig:sync-examples}
\end{table}

\section{Related Work}
\subsection{Lemmatization Resources}
In Arabic lemmatization, morphological dictionaries and analyzers serve as the primary resources for nearly all previous works in this task \cite{maamouri2010ldc,boudchiche2017alkhalil,taji2018arabic,jarrar2024alma}. These analyzers or dictionaries extract the lemmas in an out-of-context manner based on morphosyntactic features. While they provide a strong foundation for lemmatization, their reliance on predefined linguistic rules limits adaptability to contextual variations.
In this paper, we make use of the CALIMA-S31 analyzer \cite{taji2018arabic}, which extends 
SAMA31 \cite{maamouri2010ldc}, and is used inside CAMeL~Tools\cite{obeid-etal-2020-camel}.

Several benchmark datasets exist for Arabic lemmatization, including the Penn Arabic Treebank \cite{maamouri2004penn}, Zaebuc \cite{habash-palfreyman-2022-zaebuc}, Wiki News \cite{mubarak-2018-build}, Salma \cite{jarrar2024alma}, Quranic \cite{dukes2010morphological}, and Nemlar \cite{yaseen2006building}. However, most are heavily skewed toward the news genre, limiting their applicability to diverse linguistic contexts. In addition, inconsistencies in lemma definitions and diacritic conventions complicate fair comparisons across systems \cite{elgamal2024arabic}. Table~\ref{fig:sync-examples} highlights some of the differences using three example lemmas.
To address this, we apply a synchronization method to standardize lemma and diacritic representations, enabling more consistent evaluation. We also introduce a new multi-genre benchmark dataset to expand coverage beyond news and support more comprehensive assessment of lemmatization approaches. Most of the aforementioned datasets are included in our evaluation to ensure broad generalization.

\subsection{Lemmatization Approaches}

Lemmatization has been tackled through various paradigms. One common approach relies on morphological dictionaries, framing lemmatization as the selection of the correct lemma from a predefined lexicon \cite{jarrar2024alma, mubarak2017build, jongejan2009automatic, ingason2008mixed, ingolfsdottir2019nefnir}. These methods use morphosyntactic features and 
heuristics, but often fail to generalize well in contextually diverse settings.

Other studies treat lemmatization as a language modeling task, predicting lemmas and associated features based on morphological analysis \cite{MADAMIRA2014, obeid2022camelira, lagus2021learning}. While these models leverage rich linguistic resources, they may struggle with out-of-vocabulary (OOV) forms and the complexities of highly inflected languages.

A third line of work frames lemmatization as a tagging task. \cite{gesmundo-samardzic-2012-lemmatisation} model it as paradigm-based tagging, learning transformation rules over affixes instead of mapping directly to lemmas. This enables better generalization and efficient use of context. \cite{muller2024joint} extend this idea with LEMMING, a joint log-linear model that simultaneously learns lemmatization and POS tagging, showing that the two tasks benefit from being learned together.

Sequence-to-sequence (seq2seq) models represent another family of approaches. \newcite{bergmanis2018context} frame lemmatization as character-level translation, using an encoder-decoder architecture with context markers, based on the Nematus toolkit \cite{sennrich2016linguistic}. \newcite{kondratyuk2018lemmatag} employ an autoregressive decoder with Luong attention and integrate POS and sentence context features. Other recent work further explores this direction using neural seq2seq models \cite{sahala2024neural}.

This paper explores leveraging external language signals for disambiguation, reframing lemmatization as LPG (Lemma-POS-Gloss) classification rather than lemma-only prediction. We introduce a semantic cluster formulation to better handle LPG complexity.

\section{Data}
\label{data}

We report results using six existing datasets with lemmatization annotations: \textbf{ATB} \cite{maamouri2004penn}, \textbf{Nemlar} \cite{yaseen2006building}, \textbf{Quran} Corpus \cite{dukes2010morphological}, \textbf{Wiki News} \cite{mubarak-2018-build}, \textbf{ZAEBUC} \cite{habash-palfreyman-2022-zaebuc}, 
and annotate  a new data set from the \textbf{BAREC} corpus \cite{habash2024guidelines}.

\subsection{Data Preparation and Synchronization} 

As mentioned earlier, previous research on Arabic lemmatization has shown inconsistencies in both task definition and lemma representation, particularly in diacritization. 
Table~\ref{fig:sync-examples} highlights some of the differences using three example lemmas.
To address this, we align all datasets with CALIMA-S31 standards, which is based on the LDC Standard Arabic Morphological Analyzer (SAMA3.1) 
\cite{Graff:2009:standard}, and extended by \cite{taji-etal-2018-arabic}. The process involves ranking and selecting the closest LPG set for each word in a given dataset after applying normalization, and computing a synchronization score to determine the best-matching reference.  

\paragraph{Rationale for CALIMA-S31 Alignment}  
CALIMA-S31 follows the same lemma annotation and diacritization rules as the LDC. Since LDC's data (Arabic Treebank) constitutes a major linguistic resource, aligning with CALIMA-S31 ensures consistency across datasets. Also,   the CALIMA-S31 morphological database is supported by the Camel~Tools toolkit for Arabic NLP \cite{obeid-etal-2020-camel}.

\paragraph{Data LPG Synchronization Pipeline}  
For each word, we retrieve all possible LPG sets from CALIMA-S31 and rank them based on a synchronization score. The LPG set with the highest score is selected as the gold reference. If multiple candidates achieve the same highest score, a backoff strategy resolves the ambiguity.  
To ensure consistency between the gold reference and CALIMA-S31 outputs, we apply several normalization steps addressing diacritic and orthographic variations. 
A detailed list is provided in Appendix~\ref{appendix:data_prep}.

\paragraph{Calculation of Synchronization Score}  
After retrieving LPG sets and applying normalization, we compute a synchronization score across each three LPG dimensions to determine the best-matching reference. All scores are normalized to fall within a range of 0 to 1. The score computation depends on the available data dimensions, i.e.  LPG , LP , or  L, as well as on the presence of actual gold reference in the original data. 
The final choice is based on the highest synchronization score. A detailed explanation of score computation is in Appendix~\ref{appendix:data_prep}.

\begin{table}[t]
\centering
\small
\begin{tabular}{lrrr}
\toprule
\textbf{Dataset} & \textbf{All Tokens} & \textbf{News} & \textbf{Evaluatable} \\
\midrule
ATB Train   & 503{,}015  & 100.0\% & 99.1\% \\
ATB Dev     & 63{,}137   & 100.0\% & 99.2\% \\
ATB Test    & 63{,}172   & 100.0\% & 99.0\% \\
BAREC       & 98{,}676   & 18.5\%  & 96.9\% \\
Nemlar      & 480{,}417  & 52.6\%  & 98.4\% \\
Quran       & 77{,}429   & 0.0\%   & 100.0\% \\
WikiNews    & 18{,}300   & 100.0\% & 100.0\% \\
ZAEBUC      & 34{,}235   & 0.0\%   & 100.0\% \\
\midrule
\textbf{Total} & 1{,}338{,}381 & 68.6\%  & 98.8\% \\
\bottomrule
\end{tabular}
\caption{Dataset statistics: total token count, proportion from news text, and proportion with a gold lemma reference.}
\label{tab:dataset_stats}
\end{table}

\hide{
\begin{table}[t]
\centering
\small
\begin{tabular}{lrrr}
\toprule
\textbf{Dataset} & \textbf{Evaluatable} & \textbf{Non-Evaluatable} & \textbf{Total} \\
\midrule
ATB\_Train   & 498,430  & 4,585   & 503,015 \\
ATB\_Dev     & 62,609   & 528     & 63,137 \\
ATB\_Test    & 62,522   & 650     & 63,172 \\
BAREC        & 95,627   & 3,049   & 98,676 \\
Nemlar       & 472,829  & 7,588   & 480,417 \\
Quran        & 77,429   & -       & 77,429 \\
WikiNews     & 18,300   & -       & 18,300 \\
ZAEBUC       & 34,235   & -       & 34,235 \\
\midrule
\textbf{Total} & \textbf{1,321,981} & \textbf{16,400} & \textbf{1,338,381} \\
\bottomrule
\end{tabular}
\caption{Breakdown of evaluatable and non-evaluatable word tokens across datasets.}
\label{tab:evaluation_coverage}
\end{table}
\begin{table}[t]
\centering
\small
\begin{tabular}{lrrr}
\toprule
\textbf{Dataset} & \textbf{News} & \textbf{Non-News} & \textbf{Total} \\
\midrule
ATB\_Train   & 503,015  & -        & 503,015 \\
ATB\_Dev     & 63,137   & -        & 63,137 \\
ATB\_Test    & 63,172   & -        & 63,172 \\
BAREC        & 18,233   & 80,443   & 98,676 \\
Nemlar       & 252,711  & 227,706  & 480,417 \\
Quran        & -        & 77,429   & 77,429 \\
WikiNews     & 18,300   & -        & 18,300 \\
ZAEBUC       & -        & 34,235   & 34,235 \\
\midrule
\textbf{Total} & \textbf{918,568} & \textbf{419,813} & \textbf{1,338,381} \\
\bottomrule
\end{tabular}
\caption{Distribution of word tokens across datasets, categorized by news and non-news domains.}
\label{tab:news_nonnews_distribution}
\end{table}
}


As shown in Table~\ref{tab:dataset_stats}, following the synchronization stage, each dataset is either fully or partially evaluatable. In some cases, portions remain non-evaluatable due to missing gold references in the original source, preventing complete alignment. This distinction ensures that only consistently annotated data is used for evaluation, supporting fair and reliable comparisons across datasets.

\subsection{Datasets}
We evaluate our approaches across the listed datasets. As shown in Table~\ref{tab:dataset_stats}, news data accounts for nearly twice as much as all other genres combined, and our baseline disambiguator is trained exclusively on news text (ATB Train).

\label{rel-work}


We also introduce a new benchmark dataset  based on a portion of the publicly available BAREC Corpus \cite{habash2024guidelines}.  
The \textbf{BAREC Lemmatization Dataset} comprises  diverse genres like 1001 Nights, Poetry, Novels, Emarati Curriclum, ChatGPT, Subtitles, Sahih al-Bukhari and others (See Table~\ref{tab:baerc-categories} in Appendix~\ref{app:barec-nemlar}). We annotated this dataset following the standard lemmatization guidelines used in \newcite{maamouri2010ldc}, and included  the lemma, POS, and gloss for each word using CAMeL Morph MSA, an open-source morphological database with very high coverage that goes beyond CALIMA-S31 \cite{khairallah2024camel}. 
The annotation was completed by one Arabic native speaker with extensive experience in Arabic annotation.


\begin{table*}[t]

\centering
\tabcolsep2pt
\small
\begin{tabular}{lll}
\toprule
\textbf{Methodology} & \textbf{Technique} & \textbf{Required Resources} \\
\midrule
\textbf{Sequence to Sequence} & \textbf{S2S} & Character-level Transformer Generation Model + Annotated Corpus \\
\textbf{Random Selection} & \textbf{Rand} & Morphological Analyzer + Deterministic Randomization\\
\textbf{Probabilistic Selection} & \textbf{LogP} & Morphological Analyzer + Annotated Corpus\\
\textbf{Disambiguator} & \textbf{Tagger} & Morphological Analyzer + Annotated Corpus + Tagger \\
\textbf{Gloss Cosine Similarity} & \textbf{SimG} & Morphological Analyzer + Machine Translation + Sim Align + Sent Similarity LM\\
\textbf{Classification}           & \textbf{LexC} & Classifier + Annotated Corpus\\
                         & \textbf{LexC+Tagger} & Morphological Analyzer + Annotated Corpus  + Classifier \\
\textbf{Clustering} & \textbf{Clust} & Morphological Analyzer  + Annotated Corpus + Clustering Model \\
\bottomrule
\end{tabular}

    \caption{A summary of all the approaches and techniques used in this study, along with the required resources needed for implementation in any language. }
    \label{fig:Approaches_needed_resources}
\end{table*}
\section{Approach}
\label{approach}

We investigate a range of approaches with varying reliance on existing lemmatization resources, primarily morphological analyzers and annotated corpora.\footnote{While one can distinguish between out-of-context analyzers and in-context annotated corpora as different types of artifacts, we note that most annotated datasets depend on analyzer lexicons to support the manual annotation process.} Table~\ref{fig:Approaches_needed_resources} summarizes the approaches and techniques explored in this study.

Our main classification approaches start from a set of LPG candidates per word produced by a morphological analyzer, either unranked (All) or ranked by a POS tagger (Top). A classifier selects among these candidates. We also evaluate classification without an analyzer, using only the annotated corpus. We explore different classifier types that leverage various input features and model architectures.
Additionally, we test a seq2seq model that directly predicts the lemma from the input word and its context, without relying on analyzer-generated options. Finally, we investigate hybrid models that combine these techniques.

We discuss the various approaches next.

\paragraph{Sequence to Sequence model (S2S)} We trained a sequence-to-sequence model from scratch using the ATB training data. The input to the model consists of the target word along with a context window of two words before and two words after, while the output is the corresponding lemma for the target word. This setup enables the model to learn contextual patterns that inform lemma generation without relying on predefined candidate sets. Details are in Section~\ref{eval}.

\paragraph{Random Selection (Rand)} As a simple baseline, we select an LPG candidate randomly using a deterministic method: the word's index modulo the number of candidates.

\paragraph{Probabilistic Selection (LogP)}
In this approach, the system retrieves all possible LPG candidates and ranks them based solely on the log probability of the lemma and POS combination. The top-ranked candidate is selected as the final output.

\paragraph{Disambiguator (Tagger)}
This method extends probabilistic selection by first ranking LPG candidates using POS tagger scores, then sorting by lemma and POS log probabilities from the annotated corpus. The top candidate is chosen, serving as our main probabilistic baseline.

\paragraph{Gloss Cosine Similarity (SimG)} In this approach, re-ranking is based on the cosine similarity between each gloss in the LPG set and its aligned English counterpart from the translated sentence. The translation is generated using the Google Translate API, and word alignment is performed using the SimAlign RoBERTa model with the `mwmf' alignment strategy \cite{sabet2020simalign}. Both the gloss and the aligned English word are embedded using the gte-Base English language model \cite{li2023towards}, and similarity is computed using cosine similarity. If alignment fails for a given Arabic word, the similarity is instead computed between the gloss and the entire translated sentence.
\paragraph{Classification (LexC)} In this approach, lemmatization is framed as a classification task, where each unique LPG is treated as a distinct class, resulting in approximately 18,000 target classes that the model has encountered during training. To reduce noise, digits and punctuation are grouped into a single class, while words lacking a gold reference are assigned to a separate "unknown" class. A BERT model is fine-tuned in two stages. In the first stage, the model is trained using the input word along with its context, with the corresponding LPG class as the target label. In the second stage, instances without a gold label are reassigned to the most probable class based on predictions from the first model. The model is then fine-tuned again using these updated labels to further improve accuracy. The final fine-tuned model is used to select the most suitable LPG from the candidate set or to fall back on alternative selection strategies when needed.

The final fine-tuned model is utilized in two different ways: (1) directly using its prediction as the LPG (LexC), (2) checking if the predicted LPG exists in the primary LPG set from the analyzer; if it does, selecting it; otherwise, applying a fallback strategy reverting to the primary log probability-based ranking (LexC+LogP).


\paragraph{Clustering (Clust)}  In this approach, we redefine lemmatization as a clustering task, which is later transformed into a classification problem. Each unique LPG is grouped into a cluster with semantically similar entries (e.g., countries forming one cluster and cities another). Clusters are formed using a fine-tuned classification model combined with a clustering technique for known LPGs, with the number of clusters determined based on a custom evaluation metric. For unknown LPGs in the morphological database that have not yet been assigned a cluster, gloss-based cosine similarity is applied to identify the closest existing cluster, to which they are then assigned. The motivation behind this method is to reduce the search space for identifying the correct LPG from the LexC method by narrowing the candidate set to a smaller, semantically organized group. In total, we arrived at 2,000 clusters that collectively represent the entire LPG space of the analyzer. A sample of these clusters is provided in Appendix \ref{appendix:clusters_sample}.

Once the clusters are established, a classification model is fine-tuned to predict the cluster containing the correct LPG from the given primary set. If any LPGs in the primary set belong to the predicted cluster, they are extracted and re-ranked based on POS-LEX log probability, with the top-ranked option selected. If no LPGs from the primary set match the predicted cluster, the system falls back to the primary backoff technique.

The clustering process relies on a fine-tuned model to generate contextual embeddings for each word in the training set. Words that share the same LPG are assigned the same averaged embedding. These embeddings are then clustered using the K-Means algorithm. To determine the optimal number of clusters, we introduce a metric called the Cluster Compactness Ratio (CCR). This ratio is calculated as the average number of ambiguous lemmas that share the same cluster per word, divided by the total number of ambiguous lemmas. Intuitively, the goal is to minimize this value, since an ideal clustering would assign each ambiguous lemma to its own distinct cluster. By encouraging separation between competing lemmas, the CCR helps guide the selection of a cluster count that reduces ambiguity and results in tighter, semantically coherent groupings. This favors smaller, well-defined clusters over broader ones, improving the reliability of the final lemma selection process. Once the clusters are formed, a clustering model is fine-tuned specifically on the cluster labels, as previously discussed, to further refine the selection process.

Table~\ref{fig:Approaches_needed_resources} presents a  summary of all the approaches and techniques used in this study, along with the required resources needed for implementation in any language. Each method varies in computational complexity based on its dependencies. As highlighted, the most computationally expensive techniques are SimG, primarily due to their reliance on external machine translation systems, such as Google API, and alignment models that are pre-trained neural models rather than statistical ones. 
 While these methods improve disambiguation via contextual similarity, their practicality is limited by computational overhead. In contrast, classification and clustering approaches, though requiring fine-tuned models, are generally more efficient, scalable, and easier to optimize.

\begin{table*}[t]
\tabcolsep4pt
\small
\centering
\begin{tabular}{clcccccc|ccc}
\toprule
&\textbf{Technique} & \textbf{Corpus} & \textbf{Tagger} & \textbf{Analyzer} & \textbf{Classifier} & \textbf{Generator} & \textbf{Select} & \textbf{L} & \textbf{LP} & \textbf{LPG} \\
\midrule
(a) & \textbf{S2S} & L & - & - & - & S2S & - & 95.0 & - & - \\ \midrule
(b) & \textbf{LexC} & LPG & - & - & LexC & - & - & 89.5 & 88.5 & 85.6 \\
 &\textbf{LexC+S2S} & LPG & - & - & LexC & S2S & - & 95.0 & 90.0 & 74.9 \\ \midrule
(c)& \textbf{All+Rand} & - & - & AllSet & - & - & Rand & 72.9 & 64.6 & 59.4 \\
&\textbf{All+SimG} & - & - & AllSet & - & - & SimG & 91.7 & 87.0 & 83.2 \\ \midrule
(d) &\textbf{All+LogP} & LP & - & AllSet & - & - & LogP & 93.7 & 91.4 & 88.2 \\
&\textbf{All+S2S+Log}P & LP & - & AllSet & - & S2S & LogP & 97.4 & 95.0 & 91.6 \\ \midrule
(e) &\textbf{Top+Rand} & P & POS & TopSet & - & - & Rand & 93.0 & 92.3 & 87.1 \\
&\textbf{Top+SimG} & P & POS & TopSet & - & - & SimG & 98.1 & 97.3 & 94.3 \\ \midrule
(f) &\textbf{Top+LogP} & LP & POS & TopSet & - & - & LogP & 98.2 & 97.4 & 94.4 \\
&\textbf{Top+S2S+LogP} & LP & POS & TopSet & - & S2S & LogP & 98.7 & 97.9 & 94.9 \\ \midrule
(g)& \textbf{Top+LexC+LogP} & LPG & POS & TopSet & LexC & - & LogP & 98.8 & \textbf{98.1} & \textbf{95.6} \\
&\textbf{Top+LexC+S2S+LogP} & LPG & POS & TopSet & LexC & S2S & LogP & \textbf{98.9} & \textbf{98.1} & \textbf{95.6} \\ \midrule
(h) &\textbf{Top+Clust+LogP} & LPG & POS & TopSet & Clust & - & LogP & \textbf{98.9} & \textbf{98.1} & \textbf{95.6} \\
&\textbf{Top+Clust+S2S+LogP} & LPG & POS & TopSet & Clust & S2S & LogP & \textbf{98.9} & \textbf{98.1} & 95.4 \\
\bottomrule
\end{tabular}
\caption{Comparison of techniques across different configurations on the ATB Dev set. The table summarizes the components used in each setup, including the corpus type, tagger, analyzer, classifier, generator, and tiebreaking method.}
\label{tab:atb_dev_set_techniques}
\end{table*}

\section{Evaluation}
\label{eval}

\subsection{Experimental Setups}
\paragraph{Data} The data used for training the unigram log probability model and fine-tuning the classification and clustering models was derived from the ATB123 Train set, following the same splits outlined in the literature \cite{Diab:2013:ldc,khalifa2020morphological,zalmout2020morphological,inoue2021morphosyntactic} or provided by the data set creators. This ensures consistency with prior work and enables direct comparison of results across different methodologies.

\paragraph{Metrics} Results report accuracy over L, LP, or LPG matches on evaluatable data, counting all tokens.

\paragraph{Building the Sequence to Sequence Model}We trained a sequence-to-sequence model for Arabic lemmatization from scratch, using a 6-layer encoder-decoder architecture with 6 attention heads, a hidden size of 512, and a feed-forward dimension of 2048. Dropout was set to 0.2 across all components, and input sequences were capped at 64 tokens to match the context window size. The model was trained without caching and initialized with the padding token for decoding.

Training was conducted on 3 parallel NVIDIA A100 GPUs and completed in approximately 5 hours. We used Hugging Face’s \texttt{Seq2SeqTrainer} with a learning rate of $5 \times 10^{-5}$, batch sizes of 64 (train) and 32 (eval), and 100 epochs. Gradient checkpointing and FP16 precision were enabled to optimize memory and speed. The best model was selected based on validation accuracy evaluated at the end of each epoch.

\paragraph{Fine-Tuning for Classification and Clustering} The CAMeL BERT msa\_pos\_MSA model \cite{inoue2021interplay} is fine-tuned for both classification and clustering tasks. Training is performed over 10 epochs with a learning rate of $2 \times 10^{-5}$, a batch size of 16, and a maximum sequence length of 512. Three fine-tuned models were trained on an A100 GPU, with an estimated training time of 60 minutes per model.

\paragraph{Disambiguator \& Morphological Analyzer DB} All experiments use the CAMeL-unfactored BERT disambiguator model as the baseline competitor \cite{inoue2021morphosyntactic}, with CALIMA-S31 \cite{taji2018arabic} as the morphological analyzer DB and NOAN\_PROP as the backoff technique, as implemented in CAMEL Tools \cite{obeid-etal-2020-camel}.\footnote{CamelTools v1.5.5: Bert-Disambig\textbf{+}calima-msa-s31 db.}

\begin{table*}[ht]
\centering
\small
\tabcolsep4pt
\begin{tabular}{l|ccc|ccc|c|cc|cc|ccc}
\toprule
\textbf{Dataset} & \multicolumn{3}{c|}{\textbf{ATB Test}} & \multicolumn{3}{c|}{\textbf{Barec}} & \textbf{Nemlar} & \multicolumn{2}{c|}{\textbf{Quran}} & \multicolumn{2}{c|}{\textbf{WikiNews}} & \multicolumn{3}{c}{\textbf{Zaebuc}} \\
\textbf{Tag}     & \textbf{L }& \textbf{LP} & \textbf{LPG} & \textbf{L} & \textbf{LP} & \textbf{LPG} & \textbf{L}      & \textbf{L} & \textbf{LP} & \textbf{L} & \textbf{LP} & \textbf{L} & \textbf{LP} & \textbf{LPG} \\
\midrule

\textbf{S2S }& 95.0 & - & - & 87.0 & - & - & 83.6 & 65.7 & - & 90.5 & - & 92.5 & - & - \\\midrule
\textbf{LexC+S2S} & 95.0 & 90.4 & 75.0 & 87.0 & 78.0 & 64.2 & 83.6 & 65.7 & 61.7 & 90.5 & 86.9 & 92.5 & 90.6 & 77.1 \\\midrule
\textbf{All+Rand} & 73.1 & 64.7 & 59.8 & 69.9 & 62.7 & 57.7 & 62.4 & 55.3 & 46.6 & 68.6 & 61.1 & 67.0 & 61.6 & 57.2 \\
\textbf{Top+Rand} & 92.9 & 92.2 & 87.3 & 90.2 & 89.1 & 83.9 & 84.0 & 77.9 & 75.8 & 89.1 & 87.8 & 90.8 & 89.6 & 84.5 \\\midrule
\textbf{Top+LogP }& 98.0 & 97.3 & 94.6 & 96.4 & 95.3 & 92.1 & 89.7 & 83.4 & 81.3 & 94.5 & 93.1 & 96.2 & 95.0 & 91.2 \\
\textbf{Top+S2S+LogP} & 98.6 & 97.9 & 94.6 & 96.7 & 95.5 & 92.4 & 90.3 & 83.6 & 81.4 & 94.9 & 93.5 & 96.9 & 95.7 & 91.9 \\\midrule
\textbf{Top+LexC+LogP}& 98.7 & 98.0 & 95.9 & \textbf{97.1} & \textbf{96.0} & \textbf{92.7} & \textbf{90.5} & \textbf{84.8} & \textbf{82.6} & 95.1 & 93.7 & 97.3 & 96.1 & 92.3 \\
\textbf{Top+LexC+S2S+LogP}& 98.7 & \textbf{98.1} & \textbf{96.0} & 97.0 & 95.9 & 92.6 & \textbf{90.5} & 84.7 & \textbf{82.6} & 95.1 & \textbf{93.8} & 97.3 & 96.1 & 92.3 \\\midrule
\textbf{Top+Clust+LogP} & 98.7 & 98.0 & 95.5 & \textbf{97.1} & \textbf{96.0} & 92.6 & \textbf{90.5} & 84.6 & 82.5 & \textbf{95.2} & \textbf{93.8} & \textbf{97.5} & \textbf{96.3} & \textbf{92.4} \\
\textbf{Top+Clust+S2S+LogP} & \textbf{98.8} & \textbf{98.1} & 95.7 & 97.0 & 95.9 & 92.6 & \textbf{90.5} & 84.2 & 82.1 & \textbf{95.2} & \textbf{93.8} & 97.3 & 96.1 & 92.2 \\
\bottomrule
\end{tabular}
\caption{Performance of different systems evaluated on multiple test sets across varying tag set granularities.}
\label{tab:test_sets_experiments}
\end{table*}


We first evaluated all our approaches on the ATB123 dev set, as presented in Table~\ref{tab:atb_dev_set_techniques}. These approaches were assessed using three different evaluation granularities, as previously mentioned. Our experiments on the dev set were conducted in a variety of configurations, each representing a distinct combination of the proposed techniques. For each configuration, we considered multiple factors: whether the technique operates independently or depends on prior annotation, whether it relies on an external tagger, whether it incorporates the morphological analyzer, whether it has access to the full set of LPG candidates or only the top-ranked option, whether it integrates outputs from the classification or clustering models. whether the technique includes a generator (e.g., seq2seq model) and how tie-breaking is handled when multiple candidates remain.

Each technique or combination of techniques is treated as a sequential pipeline. For example, the setup “Top+Clust+S2S+LogP” follows a sequential process: retrieve the top-ranked LPG set, filter by predicted cluster, match the seq2seq-predicted lemma, and finally select the candidate with the highest log probability. This modular evaluation framework allows us to compare the contribution of each component under controlled conditions.

\subsection{Results}
Results in Table~\ref{tab:atb_dev_set_techniques} highlight several insights about the performance of different lemmatization strategies. In group (a), the seq2seq model trained independently of the analyzer achieves strong results (95.0\% accuracy), outperforming the LexC classifier in group (b), even when combined with seq2seq. This reflects the benefit of a generative model that is not limited by a predefined candidate set.

Group (c) focuses on setups using only the analyzer. The random selection method (All+Rand) performs poorly, but adding gloss-based similarity (All+SimG) significantly improves results, demonstrating the usefulness of semantic signals in the absence of other models.

In group (d), introducing lemma and POS information (LP) through log probability ranking improves performance. Adding the seq2seq model further boosts accuracy by helping narrow down the correct lemma more precisely.

Groups (e) and (f) evaluate scenarios with access to POS tags and only the top-ranked candidates. These represent practical, efficient setups. The “Top+LogP” method provides a strong baseline, and using the seq2seq model as a filter (Top+S2S+LogP) improves it even further.

Finally, groups (g) and (h) incorporate richer supervision, LexC classification or LPG clustering. Both outperform the baseline, with clustering yielding better results. Adding the seq2seq model to these pipelines gives a small but consistent improvement, showing its value even in already strong configurations.

In Table~\ref{tab:test_sets_experiments}, the results across the various test sets are largely consistent with the patterns observed on the ATB123 dev set, reinforcing the generalizability and robustness of our proposed methods. Notably, the clustering-based approach demonstrates superior performance across all datasets for the lemma granularity , consistently outperforming the classification-based LexC method. This highlights the strength of semantically informed clustering in capturing lexical variation and guiding lemma selection, even in diverse and unseen domains. However, for other granularities, the two methods show competitive performance against each other. 

We measure statistical significance using the McNemar Test \cite{McNemar1947} in the highest granularity available for each test set in the All token setup.
All Top+Clust+LogP and Top+LexC+S2S+LogP improvements over Top+LogP are statistically significant ($p<0.05$).  All differences between Top+Clust+LogP and Top+LexC+S2S+LogP are statistically significant ($p<0.05$) except for wiki data and barec.

This performance difference may be attributed to the fact that the clustering technique considers and leverages information from the entire 49K LPG entries in the CALIMA-S31 database, whereas the classification-based approach is limited to approximately 18K unique classes. By incorporating a broader range of lexical knowledge, clustering may offer a more comprehensive representation, contributing to its advantage in certain datasets.

\subsection{Error Analysis}
 
We conducted a manual error analysis to better understand the failure cases of our best-performing system (\textbf{BEST}) compared to the character-level sequence-to-sequence model (\textbf{S2S}) on the ATB dev set.
Out of 62,609 evaluatable entries, S2S made 3,108 errors, BEST made 708, with 631 errors overlapping (20\% of S2S, 89\% of BEST). 
To gain insight, we randomly sampled 100 errors from the \textbf{S2S~only} set, 100 from the \textbf{S2S+BEST} overlap, and included all 77 \textbf{BEST~only} errors. Below, we report on 200 S2S and 177 BEST errors (Table~\ref{tab:error-types-simple}).
We categorized errors into three types:
\textbf{(a)~Hallucination:}  The predicted lemma is not morphologically plausible, e.g.,
  the word \<ترحيبًا>~\textit{trHybA} (reference lemma \<تَرْحِيب>~\textit{taroHiyb} \textit{`welcome'}) was lemmatized by S2S as \<تَحَرِّي> \textit{taHar{\SHADDA}iy} \textit{`investigation'};
\textbf{(b)~Plausible:} The predicted lemma is morphologically valid but differs noticeably from the reference, the word 
  \<لزهرة>~\textit{lzhrp} (reference lemma \<زَهْر>~\textit{zahor} `flower') is lemmatized as \<زُهْرَة>~\textit{zuhorap} `Venus'; and
 \textbf{(c)~Diacritization/Hamzation:} The predicted lemma differs from the reference primarily by diacritics or hamza placement, e.g., the word
  \<وتحولها>~\textit{wtHwlhA} (reference lemma\<تَحَوُّل>~\textit{taHaw{\SHADDA}ul} `change~[noun]') is lemmatized as \<تَحَوَّل>~{taHaw{\SHADDA}al} `change~[verb]'. 

 S2S often hallucinated implausible lemmas (40\%), while BEST showed no hallucinations and mostly subtle diacritic or variant errors, indicating classification methods produce more morphologically consistent lemmas for Arabic.

\begin{table}[t]
\centering
\small 
\begin{tabular}{lcc}
\toprule
\textbf{Error Type} & \textbf{S2S} & \textbf{BEST} \\
\midrule
\textbf{Hallucination}       & 40.0\% & 0.0\%  \\
\textbf{Plausible}            & 26.5\% & 52.5\% \\
\textbf{Diacritization/Hamzation} & 33.5\% & 47.5\% \\

\bottomrule
\end{tabular}
\caption{Error type distribution for S2S and BEST systems.}
\label{tab:error-types-simple}
\end{table}

\hide{ 
\begin{table}[t]
    \centering
    \includegraphics[width=\linewidth]{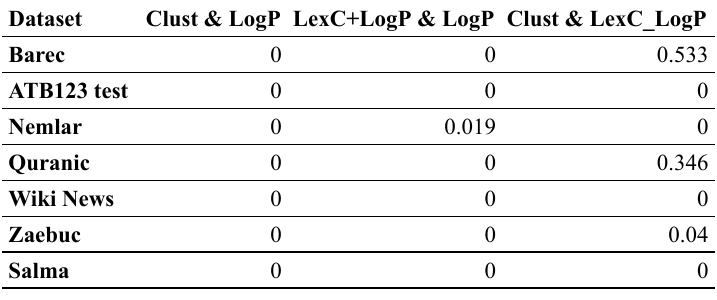}
    \caption{Significance Analysis Across All Test Sets}
    \label{fig:significance Testing analysis}
\end{table}

As shown in Table~\ref{fig:significance Testing analysis}, the significance analysis was performed across all test sets, focusing on the highest level of ambiguity in each. The results indicate statistically significant differences across all datasets when comparing Clust or LexC+LogP to the baseline. However, in BAREC, the comparison between Clust and LexC+LogP resulted in a p-value of 0.533, suggesting no substantial difference between these two approaches. In contrast, for all other datasets, p-values remained below 0.05, confirming that the observed performance differences are statistically meaningful. This highlights the effectiveness of each technique in handling high ambiguity cases.
 
}




 
\section{Conclusion and Future Work}

We introduced new lemmatization methods by framing the task as classification and clustering in the Lemma-POS-Gloss (LPG) space. Evaluated across multiple Arabic datasets (with synchronized benchmarks for consistency) and compared to character-level seq2seq models, our approaches showed strong cross-genre generalization and added-value hybridization. Our models also avoided the hallucination issues seen in seq2seq outputs. Significance testing confirmed that all performance gains were statistically meaningful.

We will release all annotations, synchronizations, and code to support future work. Going forward, we aim to expand training data, improve analyzer recall with broader LPG candidate generation, retrain the models on more diverse corpora, and explore seq2seq as a fallback for OOV terms to further boost robustness.

\newpage

\section{Limitations}

The classification-based model is constrained by a predefined set of approximately 18,000 LPG classes, while the clustering-based model operates over 2000 LPG clusters. Both approaches face challenges with out-of-vocabulary (OOV) lemmas, as fallback strategies may fail to select the optimal lemma even when it is present in the known sets. Moreover, relying solely on the top-ranked LPG candidate from the disambiguator can reduce recall by eliminating potentially correct alternatives. As for the sequence-to-sequence (S2S) model, error analysis revealed that it can occasionally hallucinate lemmas not grounded in the input, especially in ambiguous contexts. Its performance may further improve if trained on datasets spanning a broader range of genres, allowing it to generalize better to lexical variations and domain-specific usage.

\section{Ethics Statement}

All data used in the corpus collection and curation process are sourced responsibly and legally.
The annotation process is conducted with transparency and fairness.
The Arabic native speaker annotator who helped with the BAREC lemmatization Dataset was paid fair wages for their contribution.   
We acknowledge that enabling technologies such as lemmatization can be used with malicious intent to profile people based on their lexical choices or be used to build malicious software; this is not our intention, and we discourage it.
 
We used AI writing assistance within the scope of ``Assistance
purely with the language of the paper'' described in the ACL Policy on Publication Ethics.

\bibliography{custom,camel-bib-v3,anthology}

\begin{thebibliography}{50}
\providecommand{\natexlab}[1]{#1}

\bibitem[{Abdelrahman et~al.(2021)Abdelrahman, Alotaibi, Fox, and Balci}]{abdelrahman2021otrouha}
Eman Abdelrahman, Fatimah Alotaibi, Edward~A Fox, and Osman Balci. 2021.
\newblock Otrouha: A corpus of arabic etds and a framework for automatic subject classification.
\newblock \emph{The Journal of Electronic Theses and Dissertations}, 1(1):6.

\bibitem[{Al~Khalil et~al.(2018)Al~Khalil, Saddiki, Habash, and Alfalasi}]{AlKhalil:2018:leveled}
Muhamed Al~Khalil, Hind Saddiki, Nizar Habash, and Latifa Alfalasi. 2018.
\newblock {A Leveled Reading Corpus of Modern Standard {A}rabic}.
\newblock In \emph{Proceedings of the Language Resources and Evaluation Conference (LREC)}, Miyazaki, Japan.

\bibitem[{Belkebir and Habash(2021)}]{belkebir-habash-2021-automatic}
Riadh Belkebir and Nizar Habash. 2021.
\newblock \href {https://doi.org/10.18653/v1/2021.conll-1.47} {Automatic error type annotation for {A}rabic}.
\newblock In \emph{Proceedings of the 25th Conference on Computational Natural Language Learning}, pages 596--606, Online. Association for Computational Linguistics.

\bibitem[{Bergmanis and Goldwater(2018{\natexlab{a}})}]{Bergmanis:2018:context}
Toms Bergmanis and Sharon Goldwater. 2018{\natexlab{a}}.
\newblock Context sensitive neural lemmatization with lematus.
\newblock In \emph{Proceedings of the 2018 Conference of the North American Chapter of the Association for Computational Linguistics: Human Language Technologies, Volume 1 (Long Papers)}, volume~1, pages 1391--1400.

\bibitem[{Bergmanis and Goldwater(2018{\natexlab{b}})}]{bergmanis2018context}
Toms Bergmanis and Sharon Goldwater. 2018{\natexlab{b}}.
\newblock Context sensitive neural lemmatization with lematus.
\newblock In \emph{Proceedings of the 2018 Conference of the North American Chapter of the Association for Computational Linguistics: Human Language Technologies, Volume 1 (Long Papers)}, pages 1391--1400.

\bibitem[{Boudchiche et~al.(2017)Boudchiche, Mazroui, Bebah, Lakhouaja, and Boudlal}]{boudchiche2017alkhalil}
Mohamed Boudchiche, Azzeddine Mazroui, Mohamed Ould Abdallahi~Ould Bebah, Abdelhak Lakhouaja, and Abderrahim Boudlal. 2017.
\newblock Alkhalil morpho sys 2: A robust arabic morpho-syntactic analyzer.
\newblock \emph{Journal of King Saud University-Computer and Information Sciences}, 29(2):141--146.

\bibitem[{Buckwalter(2002)}]{Buckwalter:2002:buckwalter}
Tim Buckwalter. 2002.
\newblock Buckwalter {{A}rabic} morphological analyzer version 1.0.
\newblock Linguistic Data Consortium (LDC) catalog number LDC2002L49, ISBN 1-58563-257-0.

\bibitem[{Conforti et~al.(2018)Conforti, Huck, and Fraser}]{conforti2018neural}
Costanza Conforti, Matthias Huck, and Alexander Fraser. 2018.
\newblock Neural morphological tagging of lemma sequences for machine translation.
\newblock In \emph{Proceedings of the 13th Conference of the Association for Machine Translation in the Americas (Volume 1: Research Track)}, pages 39--53.

\bibitem[{Diab et~al.(2013)Diab, Habash, Rambow, and Roth}]{Diab:2013:ldc}
Mona Diab, Nizar Habash, Owen Rambow, and Ryan Roth. 2013.
\newblock {LDC {A}rabic treebanks and associated corpora: Data divisions manual}.
\newblock \emph{arXiv preprint arXiv:1309.5652}.

\bibitem[{Dukes and Habash(2010{\natexlab{a}})}]{dukes2010morphological}
Kais Dukes and Nizar Habash. 2010{\natexlab{a}}.
\newblock Morphological annotation of quranic arabic.
\newblock In \emph{Lrec}, pages 2530--2536.

\bibitem[{Dukes and Habash(2010{\natexlab{b}})}]{Dukes:2010:morphological}
Kais Dukes and Nizar Habash. 2010{\natexlab{b}}.
\newblock {Morphological Annotation of Quranic {A}rabic}.
\newblock In \emph{Proceedings of the Language Resources and Evaluation Conference (LREC)}, Valetta, Malta.

\bibitem[{El-Shishtawy and El-Ghannam(2014)}]{el2014lemma}
Tarek El-Shishtawy and Fatma El-Ghannam. 2014.
\newblock A lemma based evaluator for semitic language text summarization systems.
\newblock \emph{arXiv preprint arXiv:1403.5596}.

\bibitem[{Elgamal et~al.(2024)Elgamal, Obeid, Kabbani, Inoue, and Habash}]{elgamal2024arabic}
Salman Elgamal, Ossama Obeid, Tameem Kabbani, Go~Inoue, and Nizar Habash. 2024.
\newblock Arabic diacritics in the wild: Exploiting opportunities for improved diacritization.
\newblock \emph{arXiv preprint arXiv:2406.05760}.

\bibitem[{Gesmundo and Samard{\v{z}}i{\'c}(2012)}]{gesmundo-samardzic-2012-lemmatisation}
Andrea Gesmundo and Tanja Samard{\v{z}}i{\'c}. 2012.
\newblock \href {https://aclanthology.org/P12-2072} {Lemmatisation as a tagging task}.
\newblock In \emph{Proceedings of the 50th Annual Meeting of the Association for Computational Linguistics (Volume 2: Short Papers)}, pages 368--372, Jeju Island, Korea. Association for Computational Linguistics.

\bibitem[{Graff et~al.(2009)Graff, Maamouri, Bouziri, Krouna, Kulick, and Buckwalter}]{Graff:2009:standard}
David Graff, Mohamed Maamouri, Basma Bouziri, Sondos Krouna, Seth Kulick, and Tim Buckwalter. 2009.
\newblock {Standard {A}rabic Morphological Analyzer (SAMA) Version 3.1}.
\newblock Linguistic Data Consortium LDC2009E73.

\bibitem[{Habash and Palfreyman(2022)}]{habash-palfreyman-2022-zaebuc}
Nizar Habash and David Palfreyman. 2022.
\newblock \href {https://aclanthology.org/2022.lrec-1.9} {{ZAEBUC}: An annotated {A}rabic-{E}nglish bilingual writer corpus}.
\newblock In \emph{Proceedings of the Thirteenth Language Resources and Evaluation Conference}, pages 79--88, Marseille, France. European Language Resources Association.

\bibitem[{Habash et~al.(2024)Habash, Taha-Thomure, Elmadani, Zeino, and Abushmaes}]{habash2024guidelines}
Nizar Habash, Hanada Taha-Thomure, Khalid~N Elmadani, Zeina Zeino, and Abdallah Abushmaes. 2024.
\newblock Guidelines for fine-grained sentence-level arabic readability annotation.
\newblock \emph{arXiv preprint arXiv:2410.08674}.

\bibitem[{Ingason et~al.(2008)Ingason, Helgad{\'o}ttir, Loftsson, and R{\"o}gnvaldsson}]{ingason2008mixed}
Anton~Karl Ingason, Sigr{\'u}n Helgad{\'o}ttir, Hrafn Loftsson, and Eir{\'\i}kur R{\"o}gnvaldsson. 2008.
\newblock A mixed method lemmatization algorithm using a hierarchy of linguistic identities (holi).
\newblock In \emph{Advances in natural language processing: 6th international conference, GoTAL 2008 Gothenburg, Sweden, August 25-27, 2008 Proceedings}, pages 205--216. Springer.

\bibitem[{Ing{\'o}lfsd{\'o}ttir et~al.(2019)Ing{\'o}lfsd{\'o}ttir, Loftsson, Da{\dh}ason, and Bjarnad{\'o}ttir}]{ingolfsdottir2019nefnir}
Svanhv{\'\i}t~Lilja Ing{\'o}lfsd{\'o}ttir, Hrafn Loftsson, J{\'o}n~Fri{\dh}rik Da{\dh}ason, and Krist{\'\i}n Bjarnad{\'o}ttir. 2019.
\newblock Nefnir: A high accuracy lemmatizer for icelandic.
\newblock \emph{arXiv preprint arXiv:1907.11907}.

\bibitem[{Inoue et~al.(2021{\natexlab{a}})Inoue, Alhafni, Baimukan, Bouamor, and Habash}]{inoue2021interplay}
Go~Inoue, Bashar Alhafni, Nurpeiis Baimukan, Houda Bouamor, and Nizar Habash. 2021{\natexlab{a}}.
\newblock The interplay of variant, size, and task type in arabic pre-trained language models.
\newblock \emph{arXiv preprint arXiv:2103.06678}.

\bibitem[{Inoue et~al.(2021{\natexlab{b}})Inoue, Khalifa, and Habash}]{inoue2021morphosyntactic}
Go~Inoue, Salam Khalifa, and Nizar Habash. 2021{\natexlab{b}}.
\newblock Morphosyntactic tagging with pre-trained language models for arabic and its dialects.
\newblock \emph{arXiv preprint arXiv:2110.06852}.

\bibitem[{Jarrar et~al.(2024)Jarrar, Akra, and Hammouda}]{jarrar2024alma}
Mustafa Jarrar, Diyam Akra, and Tymaa Hammouda. 2024.
\newblock Alma: Fast lemmatizer and pos tagger for arabic.
\newblock \emph{Procedia Computer Science}, 244:378--387.

\bibitem[{Jongejan and Dalianis(2009)}]{jongejan2009automatic}
Bart Jongejan and Hercules Dalianis. 2009.
\newblock Automatic training of lemmatization rules that handle morphological changes in pre-, in-and suffixes alike.
\newblock In \emph{Proceedings of the Joint Conference of the 47th Annual Meeting of the ACL and the 4th International Joint Conference on Natural Language Processing of the AFNLP}, pages 145--153.

\bibitem[{Khairallah et~al.(2024)Khairallah, Khalifa, Marzouk, Nassar, and Habash}]{khairallah2024camel}
Christian Khairallah, Salam Khalifa, Reham Marzouk, Mayar Nassar, and Nizar Habash. 2024.
\newblock Camel morph msa: A large-scale open-source morphological analyzer for modern standard arabic.
\newblock In \emph{Proceedings of the 2024 Joint International Conference on Computational Linguistics, Language Resources and Evaluation (LREC-COLING 2024)}, pages 2683--2691.

\bibitem[{Khalifa et~al.(2020)Khalifa, Zalmout, and Habash}]{khalifa2020morphological}
Salam Khalifa, Nasser Zalmout, and Nizar Habash. 2020.
\newblock Morphological analysis and disambiguation for gulf arabic: The interplay between resources and methods.
\newblock In \emph{Proceedings of the Twelfth Language Resources and Evaluation Conference}, pages 3895--3904.

\bibitem[{Kondratyuk et~al.(2018{\natexlab{a}})Kondratyuk, Gaven{\v{c}}iak, Straka, and Haji{\v{c}}}]{Kondratyuk:2018:lemmatag}
Daniel Kondratyuk, Tom{\'a}{\v{s}} Gaven{\v{c}}iak, Milan Straka, and Jan Haji{\v{c}}. 2018{\natexlab{a}}.
\newblock Lemmatag: Jointly tagging and lemmatizing for morphologically rich languages with brnns.
\newblock In \emph{Proceedings of the 2018 Conference on Empirical Methods in Natural Language Processing}, pages 4921--4928.

\bibitem[{Kondratyuk et~al.(2018{\natexlab{b}})Kondratyuk, Gaven{\v{c}}iak, Straka, and Haji{\v{c}}}]{kondratyuk2018lemmatag}
Daniel Kondratyuk, Tom{\'a}{\v{s}} Gaven{\v{c}}iak, Milan Straka, and Jan Haji{\v{c}}. 2018{\natexlab{b}}.
\newblock Lemmatag: jointly tagging and lemmatizing for morphologically-rich languages with brnns.
\newblock \emph{arXiv preprint arXiv:1808.03703}.

\bibitem[{Lagus and Klami(2021)}]{lagus2021learning}
Jarkko Lagus and Arto Klami. 2021.
\newblock Learning to lemmatize in the word representation space.
\newblock In \emph{Proceedings of the 23rd Nordic Conference on Computational Linguistics (NoDaLiDa)}, pages 249--258.

\bibitem[{Li et~al.(2023)Li, Zhang, Zhang, Long, Xie, and Zhang}]{li2023towards}
Zehan Li, Xin Zhang, Yanzhao Zhang, Dingkun Long, Pengjun Xie, and Meishan Zhang. 2023.
\newblock Towards general text embeddings with multi-stage contrastive learning.
\newblock \emph{arXiv preprint arXiv:2308.03281}.

\bibitem[{Liberato et~al.(2024)Liberato, Alhafni, Khalil, and Habash}]{liberato-etal-2024-strategies}
Juan Liberato, Bashar Alhafni, Muhamed Khalil, and Nizar Habash. 2024.
\newblock \href {https://doi.org/10.18653/v1/2024.arabicnlp-1.5} {Strategies for {A}rabic readability modeling}.
\newblock In \emph{Proceedings of The Second Arabic Natural Language Processing Conference}, pages 55--66, Bangkok, Thailand. Association for Computational Linguistics.

\bibitem[{Maamouri et~al.(2004)Maamouri, Bies, Buckwalter, and Mekki}]{maamouri2004penn}
Mohamed Maamouri, Ann Bies, Tim Buckwalter, and Wigdan Mekki. 2004.
\newblock The penn arabic treebank: Building a large-scale annotated arabic corpus.
\newblock In \emph{NEMLAR conference on Arabic language resources and tools}, volume~27, pages 466--467. Cairo.

\bibitem[{Maamouri et~al.(2010)Maamouri, Graff, Bouziri, Krouna, Bies, and Kulick}]{maamouri2010ldc}
Mohamed Maamouri, David Graff, Basma Bouziri, Sondos Krouna, Ann Bies, and Seth Kulick. 2010.
\newblock Ldc standard arabic morphological analyzer (sama) version 3.1.
\newblock \emph{(No Title)}.

\bibitem[{McNemar(1947)}]{McNemar1947}
Quinn McNemar. 1947.
\newblock \href {https://doi.org/10.1007/bf02295996} {Note on the sampling error of the difference between correlated proportions or percentages}.
\newblock \emph{Psychometrika}, 12(2):153--157.

\bibitem[{Mubarak(2017)}]{mubarak2017build}
Hamdy Mubarak. 2017.
\newblock Build fast and accurate lemmatization for arabic.
\newblock \emph{arXiv preprint arXiv:1710.06700}.

\bibitem[{Mubarak(2018)}]{mubarak-2018-build}
Hamdy Mubarak. 2018.
\newblock \href {https://aclanthology.org/L18-1181} {Build fast and accurate lemmatization for {A}rabic}.
\newblock In \emph{Proceedings of the Eleventh International Conference on Language Resources and Evaluation ({LREC} 2018)}, Miyazaki, Japan. European Language Resources Association (ELRA).

\bibitem[{Muller et~al.(2024)Muller, Cotterell, Fraser, and Sch{\"u}tze}]{muller2024joint}
Thomas Muller, Ryan Cotterell, Alexander Fraser, and Hinrich Sch{\"u}tze. 2024.
\newblock Joint lemmatization and morphological tagging with lemming.
\newblock \emph{arXiv preprint arXiv:2405.18308}.

\bibitem[{Obeid et~al.(2022)Obeid, Inoue, and Habash}]{obeid2022camelira}
Ossama Obeid, Go~Inoue, and Nizar Habash. 2022.
\newblock Camelira: An arabic multi-dialect morphological disambiguator.
\newblock \emph{arXiv preprint arXiv:2211.16807}.

\bibitem[{Obeid et~al.(2020)Obeid, Zalmout, Khalifa, Taji, Oudah, Alhafni, Inoue, Eryani, Erdmann, and Habash}]{obeid-etal-2020-camel}
Ossama Obeid, Nasser Zalmout, Salam Khalifa, Dima Taji, Mai Oudah, Bashar Alhafni, Go~Inoue, Fadhl Eryani, Alexander Erdmann, and Nizar Habash. 2020.
\newblock \href {https://aclanthology.org/2020.lrec-1.868} {{CAM}e{L} tools: An open source python toolkit for {A}rabic natural language processing}.
\newblock In \emph{Proceedings of the Twelfth Language Resources and Evaluation Conference}, pages 7022--7032, Marseille, France. European Language Resources Association.

\bibitem[{Pasha et~al.(2014)Pasha, Al-Badrashiny, Kholy, Eskander, Diab, Habash, Pooleery, Rambow, and Roth}]{MADAMIRA2014}
Arfath Pasha, Mohamed Al-Badrashiny, Ahmed~El Kholy, Ramy Eskander, Mona Diab, Nizar Habash, Manoj Pooleery, Owen Rambow, and Ryan Roth. 2014.
\newblock Madamira: A fast, comprehensive tool for morphological analysis and disambiguation of arabic.
\newblock In \emph{In Proceedings of LREC}.

\bibitem[{Roth et~al.(2008)Roth, Rambow, Habash, Diab, and Rudin}]{Roth:2008:arabic}
Ryan Roth, Owen Rambow, Nizar Habash, Mona Diab, and Cynthia Rudin. 2008.
\newblock {A}rabic morphological tagging, diacritization, and lemmatization using lexeme models and feature ranking.
\newblock In \emph{Proceedings of the Conference of the Association for Computational Linguistics (ACL)}, Columbus, Ohio.

\bibitem[{Sabet et~al.(2020)Sabet, Dufter, Yvon, and Sch{\"u}tze}]{sabet2020simalign}
Masoud~Jalili Sabet, Philipp Dufter, Fran{\c{c}}ois Yvon, and Hinrich Sch{\"u}tze. 2020.
\newblock Simalign: High quality word alignments without parallel training data using static and contextualized embeddings.
\newblock \emph{arXiv preprint arXiv:2004.08728}.

\bibitem[{Sahala(2024)}]{sahala2024neural}
Aleksi Sahala. 2024.
\newblock Neural lemmatization and pos-tagging models for coptic, demotic and earlier egyptian.
\newblock In \emph{Proceedings of the 1st Workshop on Machine Learning for Ancient Languages (ML4AL 2024)}, pages 87--97.

\bibitem[{Seddah et~al.(2010)Seddah, Chrupa{\l}a, {\c{C}}etino{\u{g}}lu, van Genabith, and Candito}]{seddah-etal-2010-lemmatization}
Djam{\'e} Seddah, Grzegorz Chrupa{\l}a, {\"O}zlem {\c{C}}etino{\u{g}}lu, Josef van Genabith, and Marie Candito. 2010.
\newblock \href {https://aclanthology.org/W10-1410} {Lemmatization and lexicalized statistical parsing of morphologically-rich languages: the case of {F}rench}.
\newblock In \emph{Proceedings of the {NAACL} {HLT} 2010 First Workshop on Statistical Parsing of Morphologically-Rich Languages}, pages 85--93, Los Angeles, CA, USA. Association for Computational Linguistics.

\bibitem[{Semmar et~al.(2006)Semmar, Laib, and Fluhr}]{semmar2006deep}
Nasredine Semmar, Meriama Laib, and Christian Fluhr. 2006.
\newblock A deep linguistic analysis for cross-language information retrieval.
\newblock In \emph{LREC}, pages 2507--2510.

\bibitem[{Sennrich and Haddow(2016)}]{sennrich2016linguistic}
Rico Sennrich and Barry Haddow. 2016.
\newblock Linguistic input features improve neural machine translation.
\newblock \emph{arXiv preprint arXiv:1606.02892}.

\bibitem[{Taji et~al.(2018{\natexlab{a}})Taji, Khalifa, Obeid, Eryani, and Habash}]{taji2018arabic}
Dima Taji, Salam Khalifa, Ossama Obeid, Fadhl Eryani, and Nizar Habash. 2018{\natexlab{a}}.
\newblock An arabic morphological analyzer and generator with copious features.
\newblock In \emph{Proceedings of the fifteenth workshop on computational research in phonetics, phonology, and morphology}, pages 140--150.

\bibitem[{Taji et~al.(2018{\natexlab{b}})Taji, Khalifa, Obeid, Eryani, and Habash}]{taji-etal-2018-arabic}
Dima Taji, Salam Khalifa, Ossama Obeid, Fadhl Eryani, and Nizar Habash. 2018{\natexlab{b}}.
\newblock \href {https://doi.org/10.18653/v1/W18-5816} {An {A}rabic morphological analyzer and generator with copious features}.
\newblock In \emph{Proceedings of the Fifteenth Workshop on Computational Research in Phonetics, Phonology, and Morphology}, pages 140--150, Brussels, Belgium. Association for Computational Linguistics.

\bibitem[{Yaseen et~al.(2006)Yaseen, Attia, Maegaard, Choukri, Paulsson, Haamid, Krauwer, Bendahman, Fers{\o}e, Rashwan et~al.}]{yaseen2006building}
Mustafa Yaseen, Mohammed Attia, Bente Maegaard, Khalid Choukri, Niklas Paulsson, Salah Haamid, Steven Krauwer, Chomicha Bendahman, Hanne Fers{\o}e, Mohsen~A Rashwan, et~al. 2006.
\newblock Building annotated written and spoken arabic lrs in nemlar project.
\newblock In \emph{LREC}, pages 533--538. Citeseer.

\bibitem[{Zalmout(2020)}]{zalmout2020morphological}
Nasser Zalmout. 2020.
\newblock \emph{Morphological Tagging and Disambiguation in Dialectal Arabic Using Deep Learning Architectures}.
\newblock Ph.D. thesis, New York University Tandon School of Engineering.

\bibitem[{Zalmout and Habash(2020)}]{zalmout-habash-2020-joint}
Nasser Zalmout and Nizar Habash. 2020.
\newblock \href {https://doi.org/10.18653/v1/2020.acl-main.736} {Joint diacritization, lemmatization, normalization, and fine-grained morphological tagging}.
\newblock In \emph{Proceedings of the 58th Annual Meeting of the Association for Computational Linguistics}, pages 8297--8307, Online. Association for Computational Linguistics.

\end{thebibliography}

\newpage 

\appendix

\section{BAREC and Nemlar Distributions}
\label{app:barec-nemlar}

Tables~\ref{tab:baerc-categories} and~\ref{tab:nemlar-categories} illustrate the genre distribution within the BAREC and Nemlar datasets, respectively. As shown in Table~\ref{tab:baerc-categories}, BAREC covers a diverse range of genres, contributing to the increased complexity and challenge of processing this dataset. Similarly, Table~\ref{tab:nemlar-categories} presents the distribution across various genres in Nemlar, highlighting its wide coverage and relevance for evaluating lemmatization systems across different domains.

\begin{table}[h]
    \centering
    \includegraphics[width= 0.7\linewidth]{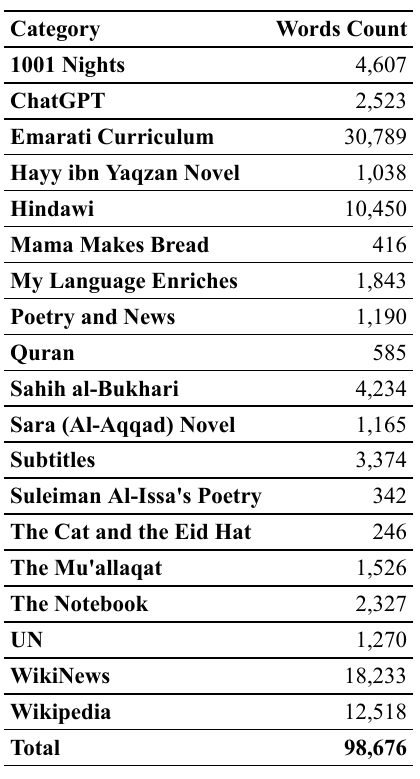}
    \caption{The distribution of various genres within the Barec dataset.}
    \label{tab:baerc-categories}
\end{table}

\begin{table}[h]
    \centering
    \includegraphics[width=0.7\linewidth]{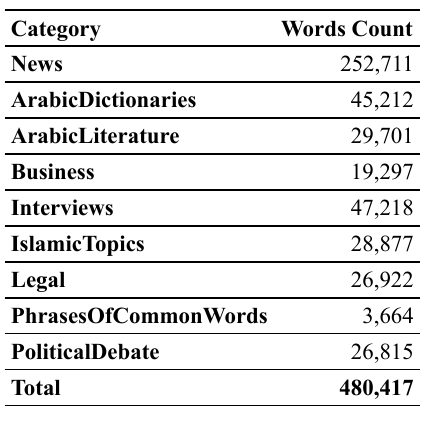}
    \caption{The distribution of all genres within the Nemlar dataset.}
    \label{tab:nemlar-categories}
\end{table}

\onecolumn
\hide{
\newpage
\section{NEW TABLES}

\begin{table}[ht]
\tabcolsep4pt
\small
\centering
\begin{tabular}{clcccccc|ccc}
\toprule
&\textbf{Technique} & \textbf{Corpus} & \textbf{Tagger} & \textbf{Analyzer} & \textbf{Classifier} & \textbf{Generator} & \textbf{Select} & \textbf{L} & \textbf{LP} & \textbf{LPG} \\
\midrule
(a) & \textbf{S2S} & L & - & - & - & S2S & - & 95.0 & - & - \\ \midrule
(b) & \textbf{LexC} & LPG & - & - & LexC & - & - & 89.5 & 88.5 & 85.6 \\
 &\textbf{LexC+S2S} & LPG & - & - & LexC & S2S & - & 94.8 & 90.0 & 74.9 \\ \midrule
(c)& \textbf{All+Rand} & - & - & AllSet & - & - & Rand & 73.3 & 65.3 & 60.0 \\
&\textbf{All+SimG} & - & - & AllSet & - & - & SimG & 91.7 & 87.0 & 83.2 \\ \midrule
(d) &\textbf{All+LogP} & LP & - & AllSet & - & - & LogP & 93.7 & 93.5 & 88.9 \\
&\textbf{All+S2S+Log}P & LP & - & AllSet & - & S2S & LogP & 97.4 & 95.0 & 91.6 \\ \midrule
(e) &\textbf{Top+Rand} & P & POS & TopSet & - & - & Rand & 93.0 & 92.3 & 87.1 \\
&\textbf{Top+SimG} & P & POS & TopSet & - & - & SimG & 98.1 & 97.3 & 94.3 \\ \midrule
(f) &\textbf{Top+LogP} & LP & POS & TopSet & - & - & LogP & 98.2 & 97.4 & 94.8 \\
&\textbf{Top+S2S+LogP} & LP & POS & TopSet & - & S2S & LogP & 98.7 & 97.9 & 94.9 \\ \midrule
(g)& \textbf{Top+LexC+LogP} & LPG & POS & TopSet & LexC & - & LogP & 98.8 & \textbf{98.1} & \textbf{95.6} \\
&\textbf{Top+LexC+S2S+LogP} & LPG & POS & TopSet & LexC & S2S & LogP & 98.7 & 97.9 & 95.4 \\ \midrule
(h) &\textbf{Top+Clust+LogP} & LPG & POS & TopSet & Clust & - & LogP & 98.8 & \textbf{98.1} & 95.4 \\
&\textbf{Top+Clust+S2S+LogP} & LPG & POS & TopSet & Clust & S2S & LogP & \textbf{98.9} & \textbf{98.1} & 95.4 \\
\bottomrule
\end{tabular}
\caption{Comparison of techniques across different configurations. \# \# \# \# \# \# \# \# \# \# \# \# \# \# \# \# \# \# \# \# \# \# \# \# \# \# \# \# \# \# \# \# \# \# \# \# \# \# }
\label{tab:techniques}
\end{table}

\begin{table*}[ht]
\centering
\small
\tabcolsep4pt
\begin{tabular}{l|ccc|ccc|c|cc|cc|ccc}
\toprule
\textbf{Dataset} & \multicolumn{3}{c|}{\textbf{ATB123 Test}} & \multicolumn{3}{c|}{\textbf{Barec}} & \textbf{Nemlar} & \multicolumn{2}{c|}{\textbf{Quran}} & \multicolumn{2}{c|}{\textbf{WikiNews}} & \multicolumn{3}{c}{\textbf{Zaebuc}} \\
\textbf{Tag}     & \textbf{L }& \textbf{LP} & \textbf{LPG} & \textbf{L} & \textbf{LP} & \textbf{LPG} & \textbf{L}      & \textbf{L} & \textbf{LP} & \textbf{L} & \textbf{LP} & \textbf{L} & \textbf{LP} & \textbf{LPG} \\
\midrule

\textbf{S2S }& 94.9 & - & - & 81.1 & - & - & X & 63.3 & - & 90.2 & - & 92.4 & - & - \\\midrule
\textbf{LexC+S2S} & 94.2 & 89.7 & 74.2 & 81.1 & 77.3 & 64.2 & 83.6 & 62.1 & 58.0 & 90.2 & 86.9 & 92.4 & 90.5 & 77.0 \\\midrule
\textbf{All+Rand} & 73.4 & 65.5 & 60.0 & 70.4 & 64.0 & 59.2 & 62.8 & 54.2 & 46.2 & 69.8 & 62.7 & 68.1 & 63.0 & 58.7 \\
\textbf{Top+Rand} & 92.9 & 92.2 & 86.9 & 90.2 & 89.1 & 83.9 & 84.0 & 78.0 & 75.9 & 89.1 & 87.8 & 90.8 & 89.6 & 84.5 \\\midrule
\textbf{Top+LogP }& 98.0 & 97.3 & 94.2 & 96.4 & 95.3 & 92.1 & 89.7 & 83.4 & 81.3 & 94.5 & 93.1 & 96.2 & 95.0 & 91.2 \\
\textbf{Top+S2S+LogP} & 98.6 & 97.9 & 94.8 & 96.6 & 95.5 & 92.4 & 89.6 & 79.9 & 77.7 & 94.9 & 93.5 & 96.9 & 95.8 & 92.0 \\
\textbf{Top+LexC+S2S+LogP }& 98.6 & 97.9 & 95.4 & 96.6 & 95.5 & 92.2 & X & 80.0 & 77.8 & 94.9 & 93.5 & 96.9 & 95.8 & 92.1 \\\midrule
\textbf{Top+Clust+LogP} & 98.7 & 98.0 & 95.1 & 97.1 & 96.0 & 92.6 & 90.5 & 81.0 & 78.9 & 95.2 & 93.8 & 97.5 & 96.4 & 92.5 \\
\textbf{Top+Clust+S2S+LogP} & 98.8 & 98.1 & 95.2 & 97.0 & 95.9 & 92.6 & 90.3 & 80.7 & 78.5 & 95.2 & 93.8 & 97.3 & 96.2 & 92.3 \\
\bottomrule
\end{tabular}
\caption{Performance of different systems across datasets and tag sets.}
\label{tab:test_sets_experiments}
\end{table*}

\begin{table}[h]
\centering
\begin{tabular}{lcccc}
\toprule
\textbf{Error Type} & \textbf{S2S only} & \multicolumn{2}{c}{\textbf{BEST+S2S}} & \textbf{BEST only}  \\
\midrule
\# of Errors         & 2477              & \multicolumn{2}{c}{631}              & 77  \\
\midrule
Diacritization  & 18\%            & 49\%            & 55\%             & 38\% \\
Plausible            & 19\%            & 34\%            & 45\%             & 62\% \\
Hallucination       & 63\%            & 17\%            & 0\%              & 0\% \\

\bottomrule
\end{tabular}
\caption{Qualitative error analysis of different error class distributions (\%) across different system configurations.
The number of errors is over the whole Dev Set; but the error types are for a sample of 100, 100, and 77, respectivly.}
\label{tab:error-analysis}
\end{table}

}

\section{License}\label{appendix:license}
In Table~\ref{table:license}, we list the license of the data and tools used in this work.
All of them are used under their intended use.

\begin{table*}[ht]
\centering
\small
\begin{tabular}{cc}
\toprule
Data/tool & License \\
\midrule
Arabic Treebank: Part 1 v 4.1 (LDC2010T13) & LDC User Agreement for Non-Members \\
Arabic Treebank: Part 2 v 3.1 (LDC2011T09) & LDC User Agreement for Non-Members \\
Arabic Treebank: Part 3 v 3.2 (LDC2010T08) & LDC User Agreement for Non-Members \\
BAREC~\cite{habash2024guidelines} &  Creative Commons
Attribution-NonCommercial-ShareAlike 4.0\\
Nemlar~\cite{yaseen2006building} &  Non Commercial Use - ELRA END USER\\
Quran~\cite{dukes2010morphological} & GNU General Public License \\
WikiNews~\cite{mubarak-2018-build} & Creative Commons Attribution 4.0 License \\
ZAEBUC~\cite{habash-palfreyman-2022-zaebuc} & Creative Commons
Attribution-NonCommercial-ShareAlike 4.0\\
CAMeL Tools~\cite{obeid-etal-2020-camel} & MIT License \\
CAMeLBERT~\cite{inoue2021morphosyntactic} & MIT License \\
\bottomrule
\end{tabular}
\caption{
License of the data and tools.
}
\label{table:license}
\end{table*}

\section{Data Preparation \& Synchronization}\label{appendix:data_prep}
This section outlines the normalization procedures and scoring criteria used during the data synchronization stage. These steps ensure consistency between CALIMA-S31 outputs and the reference annotations.

\subsection*{Normalization Procedures}
The following normalization operations were applied to address diacritic inconsistencies, orthographic variations, and dataset-specific irregularities:

\begin{itemize}  
    \item \textbf{Alef Maqsura Normalization:} Convert Alef Maqsura followed by Kasra (\<ىِ> \textit{Yi}) into Yaa followed by Kasra (\<يِ> \textit{yi}).  
    \item \textbf{Shadda Order Correction:} Ensure Shadda always precedes any associated diacritic.  
    \item \textbf{Alef Wasla Standardization:} Replace Alef Wasla with Kasra (\<ٱ> \textit{\{i}) with a normalized Alef with Kasra (\<اِ> \textit{Ai}).  
    \item \textbf{Diacritic Removal Before Long Vowels:} Remove diacritics appearing before long vowels.  
    \item \textbf{Dagger Alef Adjustment:} Remove Fatha before Dagger Alef (\textit{`}) and replace Dagger Alef with Fatha (\textit{a}).  
    \item \textbf{Tanween Positioning:} Shift Tanween to the last letter if it appears before it (e.g., \<ايضًا> \textit{AyDFA} $\rightarrow$ \<ايضاً> \textit{AyDAF}).  
    \item \textbf{Final Letter Diacritics:} Remove all diacritics on the last letter except Shadda.  
    \item \textbf{Sun Letter Shadda Removal:} Remove erroneous lemma-initial Sun Letter Shaddas, specifically in the Quran corpus \cite{Dukes:2010:morphological}.  
    \item \textbf{Alef Wasla Normalization:} Normalize Alef Wasla (\<ٱ> \textit{\{}) to Alef (\<ا>~\textit{A}).  
    \item \textbf{Dataset-Specific Adjustments:} Certain datasets required additional handling for special cases. Manual adjustments were applied where needed. All synchronization procedures will be made publicly available.
\end{itemize}

\subsection*{Scoring Criteria}
The following criteria were used to compute synchronization scores and identify the best-matching LPG set for each token:

\begin{itemize}  
    \item \textbf{Lemma Score:} Assign a score of 1 if the predicted lemma matches the gold lemma. Otherwise, compute a penalty based on edit distance.  
    \item \textbf{POS Score:} Assign 1 for a POS match and 0 for a mismatch.  
    \item \textbf{Gloss Score:} Calculate the intersection between the gold gloss and each gloss in the CALIMA-S31 output.
\end{itemize}

\section{Examples of Clusters}\label{appendix:clusters_sample}

Table~\ref{tab:clusters-sample} presents a representative sample of the lexical clusters generated automatically by the fine-tuned classification model. Each cluster groups together words that share similar semantic or functional properties, such as traits, foreign names, places, and vehicles.

\begin{table}[h]
    \centering
    \includegraphics[width= 0.9\linewidth]{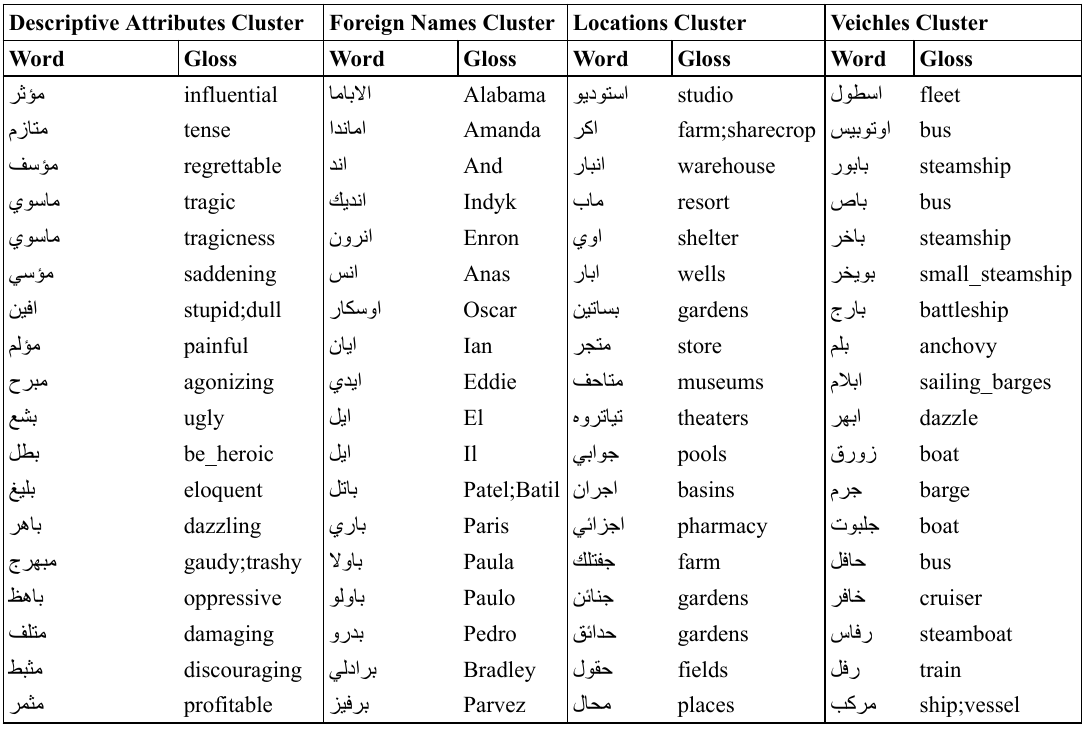}
    \caption{Examples of words grouped into semantic clusters. Each word is paired with its English gloss.}
    \label{tab:clusters-sample}
\end{table}

\clearpage
\section{S2S Lex \& Word INV/OOV Analysis}

This analysis was conducted on the ATB Dev dataset to evaluate the model’s accuracy when predicting both diacritized and undiacritized lemma forms. Since the model is trained as a character-level sequence-to-sequence system, we aimed to assess its sensitivity to surface diacritization

\begin{table}[h]
\centering
\small
\begin{tabular}{lrrr}
\toprule
\textbf{Case} & \textbf{Frequency} & \textbf{Predicted Words} & \textbf{Accuracy (\%)} \\
\midrule
\multicolumn{4}{l}{\textbf{Diacritized}} \\
\hspace{1em}Overall & 62,609 & 59,495 & 95.0 \\
\hspace{1em}(W-INV, L-INV) & 57,963 & 56,878 & 98.1 \\
\hspace{1em}(W-OOV, L-INV) & 3,722 & 2,568 & 69.0 \\
\hspace{1em}(W-INV, L-OOV) & 48 & 1 & 2.1 \\
\hspace{1em}(W-OOV, L-OOV) & 876 & 49 & 5.6 \\
\midrule
\multicolumn{4}{l}{\textbf{Undiacritized}} \\
\hspace{1em}Overall & 62,609 & 60,208 & 96.2 \\
\hspace{1em}(W-INV, L-INV) & 57,963 & 57,188 & 98.7 \\
\hspace{1em}(W-OOV, L-INV) & 3,722 & 2,684 & 72.1 \\
\hspace{1em}(W-INV, L-OOV) & 48 & 17 & 35.4 \\
\hspace{1em}(W-OOV, L-OOV) & 876 & 319 & 36.4 \\
\bottomrule
\end{tabular}
\caption{Prediction accuracy across diacritized and undiacritized inputs, broken down by in-vocabulary (INV) and out-of-vocabulary (OOV) word and lemma status.}
\label{tab:diacritic_accuracy_breakdown}
\end{table}

\section{Lemma-Level Coverage Analysis Across Datasets}
\begin{table}[h]
\centering
\small
\begin{tabular}{lrrrrrr}
\toprule
\textbf{} & \textbf{Train-INV } & \textbf{Train-INV } & \textbf{Train-OOV  } & \textbf{Train-OOV  } & \textbf{   } & \textbf{ } \\
\textbf{Dataset} & \textbf{  Analyzer-INV} & \textbf{ Analyzer-OOV} & \textbf{ Analyzer-INV} & \textbf{Analyzer-OOV} & \textbf{No Reference} & \textbf{Total} \\

\midrule
ATB\_Train    & 498,430 & 0     & 0     & 0     & 4,585   & 503,015 \\
\midrule
\multicolumn{7}{c}{\textbf{All Tests}}\\
\midrule
ATB\_Dev      & 61,740  & 0     & 869   & 0     & 528     & 63,137 \\
ATB\_Test     & 61,732  & 0     & 790   & 0     & 650     & 63,172 \\
BAREC        & 91,941  & 0     & 3,185 & 501   & 3,049   & 98,676 \\
Nemlar       & 438,203 & 0     & 14,023 & 20,603 & 7,588   & 480,417 \\
Quran        & 66,122  & 0     & 6,358 & 4,949 & 0       & 77,429 \\
WikiNews     & 17,537  & 0     & 318   & 445   & 0       & 18,300 \\
ZAEBUC       & 33,729  & 0     & 334   & 172   & 0       & 34,235 \\
\midrule
\textbf{All Tests} & \textbf{771,004} & \textbf{0} & \textbf{25,877} & \textbf{26,670} & \textbf{11,815} & \textbf{835,366} \\
\textbf{Percentage} & 92.3\% & 0.0\% & 3.1\% & 3.2\% & 1.4\% & \\
\bottomrule
\end{tabular}
\caption{Lemma-level analysis across all datasets used. This breakdown shows how many lemmas per dataset exist in the training data and/or analyzer (INV/OOV), and how many tokens have no gold reference, making them non-evaluatable.}
\label{tab:lemma_train_analyzer_analysis}
\end{table}

This table presents a lemma-level analysis across all datasets used in the study. It categorizes each token based on whether its lemma is in-vocabulary (INV) or out-of-vocabulary (OOV) with respect to both the training set and the analyzer. Cases with no gold reference are marked as non-evaluatable.

\end{document}